\documentclass[11pt]{article}

\usepackage{spconf,amsmath,epsfig}
\usepackage{amsfonts}
\usepackage{amssymb}  
\usepackage[ruled,linesnumbered]{algorithm2e} 

\usepackage{float}

\usepackage{booktabs}

\usepackage{enumitem}
\setenumerate[1]{itemsep=0pt,partopsep=0pt,parsep=\parskip,topsep=0pt}
\setitemize[1]{itemsep=0pt,partopsep=0pt,parsep=\parskip,topsep=0pt}
\setdescription{itemsep=0pt,partopsep=0pt,parsep=\parskip,topsep=0pt}

\title{Progressive Scale-aware Network for Remote Sensing Image Change Captioning}

\name{Chenyang Liu\textsuperscript{1,3}, Jiajun Yang\textsuperscript{1,3}, Zipeng Qi\textsuperscript{1,3}, Zhengxia Zou\textsuperscript{2,3}, Zhenwei Shi\textsuperscript{1,3}
\thanks{Thanks to the National Key Research and Development Program of China (Grant No. 2022ZD0160401), the National Natural Science Foundation of China under the Grants 62125102, the Beijing Natural Science Foundation under Grant JL23005, and the Fundamental Research Funds for the Central Universities. \emph{Corresponding author: Zhenwei Shi (e-mail: shizhenwei@buaa.edu.cn).}}
}
\address{\textsuperscript{1}Image Processing Center, School of Astronautics, Beihang University, Beijing 100191, China\\
\textsuperscript{2} Department of Guidance, Navigation and Control, School of Astronautics, Beihang University, \\Beijing 100191, China\\
\textsuperscript{3} Shanghai Artificial Intelligence Laboratory, Shanghai 200232, China\\
}
\begin{document}
%
\maketitle
\begin{abstract}
Remote sensing (RS) images contain numerous objects of different scales, which poses significant challenges for the RS image change captioning (RSICC) task to identify visual changes of interest in complex scenes and describe them via language. However, current methods still have some weaknesses in sufficiently extracting and utilizing multi-scale information. In this paper, we propose a progressive scale-aware network (PSNet) to address the problem. PSNet is a pure Transformer-based model. To sufficiently extract multi-scale visual features, multiple progressive difference perception (PDP) layers are stacked to progressively exploit the differencing features of bitemporal features. To sufficiently utilize the extracted multi-scale features for captioning, we propose a scale-aware reinforcement (SR) module and combine it with the Transformer decoding layer to progressively utilize the features from different PDP layers. Experiments show that our PSNet outperforms previous methods. Our code is public at \emph{{https://github.com/Chen-Yang-Liu/PSNet}} 






\end{abstract}
\begin{keywords}
Remote sensing image, change captioning, Transformer, scale-aware reinforcement
\end{keywords}
\section{Introduction}
\label{sec:intro}
Remote sensing image change captioning (RSICC) is the process of comparing bitemporal remote sensing (RS) images and describing the differences between the images in natural language. Unlike the change detection technology \cite{Shi2020, Zhu2022}, the RSICC focuses more on understanding semantic-level changes in images rather than pixel-level changes. The technology can be widely used in land planning, disaster assessment, etc \cite{applications}. 

RSICC is a recently emerging task in the RS community. Similar to the RS image captioning task \cite{wang2020wordSentence,Liu_2022}, current RSICC methods employ a common encoder-decoder structure. Chouaf \textit{et al.} first explored the RSICC task \cite{RSICC_1}. They utilized a convolutional neural network (CNN) to extract bitemporal image features and used a recurrent neural network (RNN) to generate sentences. Hoxha \textit{et al.} \cite{RSICC_2} designed two encoders and two decoders (i.e., RNNs and Support Vector Machines (SVMs)) as an extension of previous work \cite{RSICC_1}. Liu \textit{et al.} \cite{RSICCformer} proposed a large LEVIR-CC dataset for the RSICC task and a Transformer-based RSICC model, in which a dual-branch Transformer encoder with the cross-attention mechanism utilizes the differencing features to identify changes of interest and ignore irrelevant changes (e.g., illumination difference, insignificant changes). These previous works have facilitated the development of the RSICC task.

To further improve the model performance, one key is to improve the model's ability to identify visual changes of interest in complex scenes. The RS images contain objects of different scales \cite{Liu_2022}, which poses significant challenges in identifying and describing object attributes and complex relationships of changed objects. However, previous methods still have weaknesses in sufficiently extracting and utilizing multi-scale information in bitemporal RS images. 

\begin{figure*}
	\centering
	\includegraphics[width=0.75\linewidth]{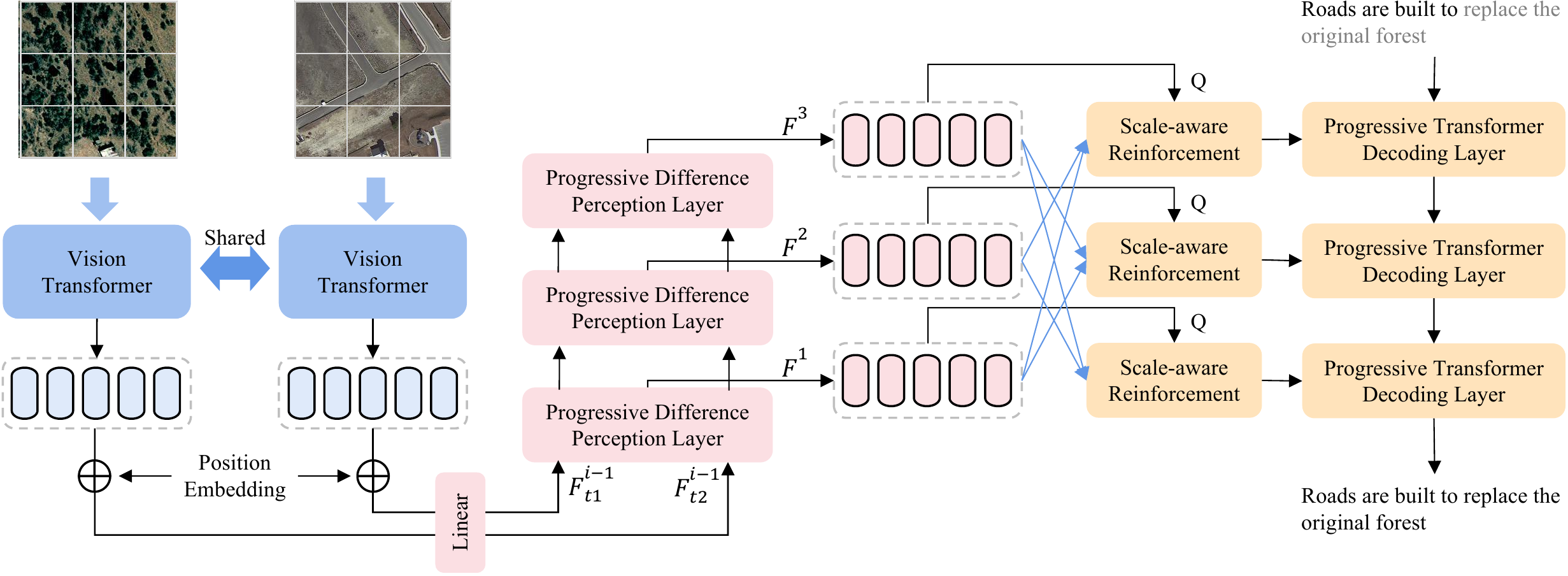}
    	 \caption{Illustration of our proposed PSNet. Our PSNet is purely Transformer based. Multiple PDP layers and SR modules are used to sufficiently extract and enhance multi-scale visual features for caption decoding.
      }
        \label{fig:overall}
\end{figure*}

In this paper, we propose a progressive scale-aware network (PSNet) to address the problem from two perspectives: extraction and utilization of multi-scale information. Our PSNet is a pure Transformer-based model. PSNet employs progressive difference perception (PDP) layers to utilize the differences between the bitemporal features to focus on the changing regions and identify visual changes. Multiple PDP layers are stacked to sufficiently extract multi-scale visual features. We propose scale-aware reinforcement (SR) module to enhance the multi-scale features from different PDP layers with multi-head cross attention \cite{Transformer}. Multiple SR modules and Transformer decoding layers are combined to sufficiently utilize the resulting multi-scale features for caption generation.



\section{The Proposed Method}
\label{sec:method}
\subsection{Overview}
Fig. \ref{fig:overall} illustrates the overall structure of our proposed PSNet. Our PSNet is purely Transformer based and consists of a vision Transformer, PDP layers, SR modules, and Transformer decoding layers. We employ the vision Transformer to extract bitemporal visual features from image pairs and add position embeddings. Multiple PDP layers progressively exploit the differences between the bitemporal features to focus on the changing regions and sufficiently extract multi-scale visual features. The extracted features from different layers are processed by multiple SR modules to improve the multi-scale feature representation capability of the model. For caption generation, we progressively send the features from each SR module into a Transformer decoding layer \cite{Transformer} to sufficiently utilize the resulting multi-scale features. This is different from the naive Transformer decoder, which uses the same features as input for each decoding layer.

\subsection{Progressive Difference Perception Layer}
\begin{figure}
	\centering
	\includegraphics[width=0.65\linewidth]{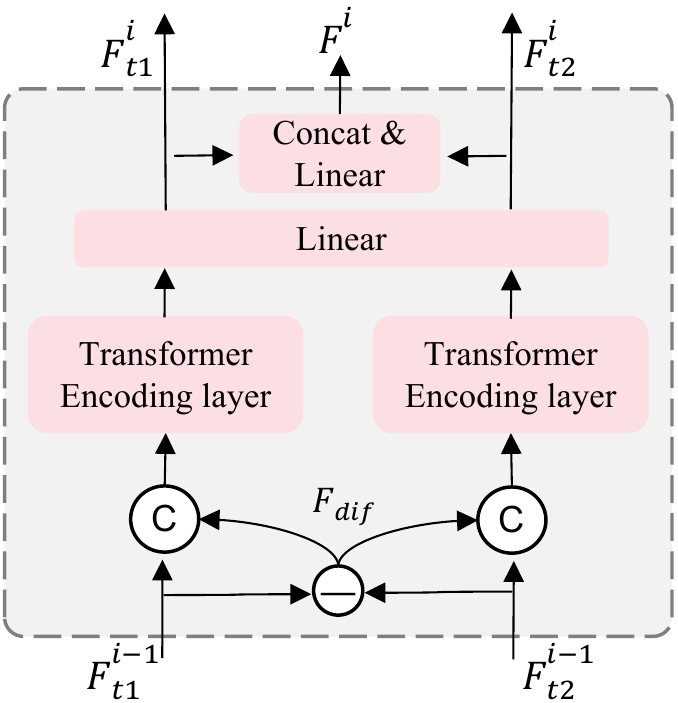}
	\caption{The PDP layer. The differencing features are used to determine changing areas and identify changes.}
	\label{fig:encoder}
\end{figure}

The structure of one PDP layer is shown in Fig. \ref{fig:encoder}. For each PDP layer, the differencing features of bitemporal features and single-temporal features are concatenated on the channel dimension. The features are then fed into two Transformer encoding layers \cite{Transformer} followed by a linear layer to focus on the changing regions and identify visual changes. The resulting bitemporal features are fed into the next PDP layer. To obtain the output of the current PDP layer, we concatenate the bitemporal features on the channel dimension and perform linear projection to fuse bitemporal features. Deep PDP layers will extract high-level semantic information, while shallow layers will retain some low-level information. We stack multiple PDP layers to sufficiently extract multi-scale visual features.

\subsection{Scale-aware Reinforcement Module}
\begin{figure}
        \centering
        \includegraphics[width=0.7\linewidth]{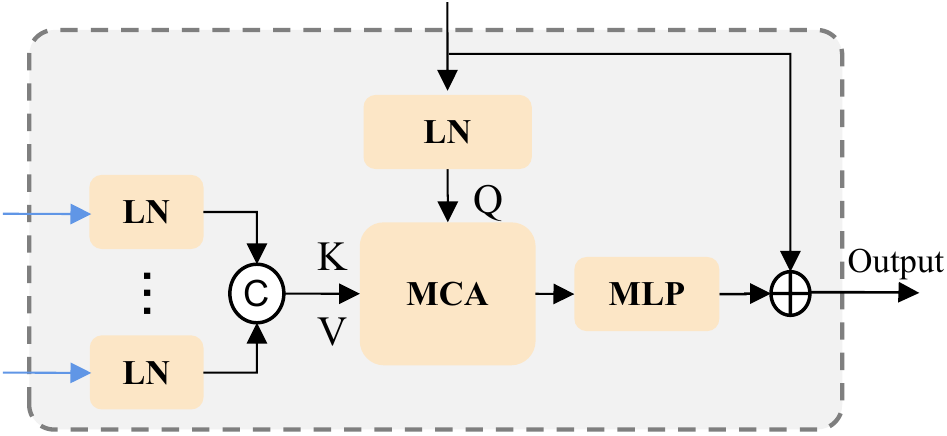}   
        \caption{The SR module. ${\rm LN}$ denotes the layer normalization \cite{LN}. ${\rm MCA}$ is the multi-head cross attention \cite{Transformer}.}
        \label{fig:Scale_aware}
\end{figure}

Instead of feeding the output of the top PDP layer to each Transformer decoding layer for caption generation, we design SR modules to process features from each PDP layer. The module aims at fusing the features of the remaining PDP layers to enhance the current features. The output features of multiple SR modules are then progressively fed into multiple Transformer decoding layers to sufficiently utilize extracted multi-scale features. Fig. \ref{fig:Scale_aware} illustrates the structure of the SR module. For example, the process of the first SR module can be formulated as follows:
\begin{align}
F_u &= [{\rm LN}(F^2);{\rm LN}(F^3)]\\
X &= {\rm MCA}({\rm LN}(F^1),F_u,F_u) \\
O_1 &= F^1+{\rm MLP}(X)
\end{align}
where $F^i(i=1,2,3)$ $\in$ $R^{N \times C}$ is the output of the $i$-th PDP layer. ${\rm LN}$ denotes the layer normalization \cite{LN} and [;] denotes the concatenation operation on the token length dimension. ${\rm MLP}$ denotes the Multilayer Perceptron. ${\rm MCA}$ is the multi-head cross attention \cite{Transformer} and can be formulated as follows:
\begin{align}
{\rm MCA}(Q,K,V) = {\rm Concat}(hd_1;...;hd_h)W^o \\
hd_i = {\rm Att}(Q W^Q_i, K W^K_i, V W^V_j)\\
{\rm Att}(Q_i, K_i, V_i) = {\rm Softmax}({\frac {{Q_i}{K_i}^T} {\sqrt{d}}})V_i
\end{align}
where $W^Q_i \in R^{C \times d}, W^K_i \in R^{C \times d}, W^V_i \in R^{C \times d}$, and $W^o \in R^{hd \times C}$ are learnable parameter matrices, $d$ is the scaling factor, $h$ is the number of heads.

\section{Experimental Results}
\label{sec:experiment}

\subsection{Experimental Setup}
\textbf{Dataset.} We conduct experiments on the LEVIR-CC dataset \cite{RSICCformer} containing 10071 pairs of bitemporal images and 50385 captions. 
Each image pair has five annotated captions. The image size is $256\times 256$ pixels with a high resolution of 0.5 m/pixel.


\textbf{Implementation Details}. Our model is implemented on PyTorch and trained using a single NVIDIA RTX 3090 GPU. The vision Transformer employs the ViT-B/32 \cite{Vit} to extract visual features with dimension 768. The PDP layers, SR modules, and Transformer decoding layers are each set to 3. The MLP of the SR module consists of two linear layers. We use the cross-entropy loss function and minimize it to optimize the model by the Adam optimizer with an initial learning rate of 0.0001. The maximum training epoch is 40.


\textbf{Evaluation Metrics.}
Following \cite{Avg_metric}, we use BLEU-N (N=1,2,3,4), ROUGE$_L$, METEOR, CIDEr, and an average metric $S^*_m$ to evaluate the quality of the generated captions. A higher metric score indicates a higher sentence accuracy.

\subsection{Experiment Results}
\label{ssec:comparison}
Table \ref{tab:Comparisons_other_methods} reports ablation experiments and the comparison results of our method and other change captioning methods on the LEVIR-CC dataset. The baseline uses two transformer encoders to process bitemporal features separately instead of using differencing features as in the PDP layer. Experimental results show that the PDP layer and SR module can effectively improve the model performance. Besides, different from the previous methods, our PSNet is purely Transformer based and helps address the multi-scale problem of the RSICC task. To identify the changed objects of different sizes, the multiple PDP layers progressively exploit differencing features to extract multi-scale features. To achieve accurate and detailed captioning, PSNet combines SR modules and Transformer decoding layers to sufficiently utilize the multi-scale features to progressively decode the visual features into descriptive sentences. With the above design, our PSNet achieves the best performance compared to the previous methods.

\begin{table*}
\renewcommand{\arraystretch}{1.0}
\caption{Comparisons experiments on the LEVIR-CC dataset, where the bolded results are the best.}
\label{tab:Comparisons_other_methods}
\centering
\begin{tabular}{c|c c c c c c c| c}
	\toprule
	Method & BLEU-1 & BLEU-2 & BLEU-3 & BLEU-4 & METEOR & ROUGE$_L$ & CIDEr & $S^*_m$\\
	\midrule
	{Capt-Rep-Diff \cite{robust_CC}} & 72.90 & 61.98 & 53.62 & 47.41 & 34.47 & 65.64 & 110.57 & 64.52\\
	{Capt-Att \cite{robust_CC}} & 77.64 & 67.40 & 59.24 & 53.15 & 36.58 & 69.73 & 121.22 & 70.17\\
	{Capt-Dual-Att \cite{robust_CC}} & 79.51 & 70.57 & 63.23 & 57.46 & 36.56 & 70.69 & 124.42 & 72.28\\
	{DUDA \cite{robust_CC}} & 81.44 & 72.22 & 64.24 & 57.79 & 37.15 & 71.04 & 124.32 & 72.58\\
	{MCCFormer-S \cite{MCCformer}} & 79.90 & 70.26 & 62.68 & 56.68 & 36.17 & 69.46 & 120.39 & 70.68\\
	{MCCFormer-D \cite{MCCformer}} & 80.42 & 70.87 & 62.86 & 56.38 & 37.29 & 70.32 & 124.44 & 72.11\\
	{RSICCFormer$_{c}$ \cite{RSICCformer}} & 83.09 & 74.32 & 66.66 & 60.44 & {38.76} & 72.63 & 130.00 & 75.46\\
	\midrule
 {Baseline} & 81.83 & 73.31 & 66.02 & 60.21 & 37.46 & 71.23 & 124.71 & 73.40\\
        {+ PDP} & 81.99 & 73.29 & 66.29 & 60.79 & 37.71 & 72.45 & 127.37 & 74.58\\
         {PSNet (+PDP +SR)} & \textbf{83.86} & \textbf{75.13} & \textbf{67.89} & \textbf{62.11} & \textbf{38.80} & \textbf{73.60} & \textbf{132.62} & \textbf{76.78}\\
	\bottomrule
\end{tabular}
\end{table*}

\section{Conclusion}
\label{sec:conclusion}

We proposed a PSNet for the RSICC task to sufficiently extract and utilize multi-scale information in the RS images. PSNet employs multiple PDP layers to sufficiently extract multi-scale visual features. Each PDP layer progressively exploits the differences between the bitemporal features to identify changes. The SR module is proposed to enhance the multi-scale features from different PDP layers with multi-head cross attention. Multiple SR modules and Transformer decoding layers are combined to sufficiently utilize the resulting multi-scale features for caption generation. Experiments show that our PSNet outperforms previous methods, and the PDP layer and the SR module are effective.


{
\scriptsize
\bibliographystyle{IEEEbib}
\bibliography{refs}
}
\end{document}